\documentclass[10pt,twocolumn,letterpaper]{article}

\usepackage{wacv}              

\usepackage{graphicx}
\usepackage{amsmath}
\usepackage{amssymb}
\usepackage{booktabs}

\usepackage[pagebackref,breaklinks,colorlinks]{hyperref}

\usepackage[capitalize]{cleveref}
\crefname{section}{Sec.}{Secs.}
\Crefname{section}{Section}{Sections}
\Crefname{table}{Table}{Tables}
\crefname{table}{Tab.}{Tabs.}

\def\wacvPaperID{19} 
\def\confName{WACV}
\def\confYear{2024}

\begin{document}

\title{Categorizing the Visual Environment and Analyzing the Visual Attention of Dogs}

\author{Shreyas Sundara Raman\\
Brown University\\
Department of Computer Science\\
\and
Madeline H. Pelgrim\\
Brown University\\
Department of CLPS\\
\and
Daphna Buchsbaum\\
Brown University\\
Department of CLPS\\
\and
Thomas Serre\\
Brown University\\
Department of CLPS\\
}

\maketitle

\begin{abstract}
   Dogs have a unique evolutionary relationship with humans and serve many important roles e.g. search and rescue, blind assistance, emotional support. However, few datasets exist to categorize visual features and objects available to dogs, as well as how dogs direct their visual attention within their environment. We collect and study a dataset with over 11,698 gazes to categorize the objects available to be gazed at by 11 dogs in everyday outdoor environments i.e. a walk around a college campus and urban area. We explore the availability of these object categories and the dogs’ visual attention over these categories using a head-mounted eye-tracking apparatus. A small portion ($~600$ images or $<20\%$ of total dataset) of the collected data is used to finetune a MaskRCNN for the novel image domain to segment objects present in the scene – enabling further statistical analysis on the visual gaze tendencies of dogs. The MaskRCNN, with eye-tracking apparatus, serves as an end-to-end model for automatically classifying the visual fixations of dogs. The fine-tuned MaskRCNN performs far better than chance; there are few individual differences between the 11 dogs and we observe greater visual fixations on buses, plants, pavement, and construction equipment. This work takes a step towards understanding visual behavior of dogs and their interaction with the physical world.
\end{abstract}

\section{Introduction}
\label{sec:intro}
Research in egocentric vision attempts to capture the visual information available to an individual, such as objects present in their environment, from a first-person perspective, as well as how they allocate their attention to those objects. This approach has become widely used to explore how human infants’ visual environments and interactions change over the course of development, and the impacts on their ability to recognize faces or acquire language. Several previous works have explored visual attention mechanisms \cite{hassanin2022visual} and visual behavior of humans \cite{Criado-Boado2019} and \cite{Chen2022}. However, little to no significant research into real-world egocentric vision and allocation of visual attention outside of humans \cite{Rodin2021} \cite{Smith2015} – particularly related to the attention mechanisms that dogs employ when perceiving and interacting in the real world. Our work attempts to understand the semantic decomposition of the domestic dog’s environment, from an egocentric perspective and then analyze how dogs direct their attention within that environment. To do this, we utilize a head-mounted eye-tracker to record objects in a dogs’ field of view together with a fine-tuned MaskRCNN for segmentation of the novel semantic classes and image distribution in our collected dataset.

The main contributions of our work include: integrating a head-mounted eye-tracker with a MaskRCNN for automatic semantic segmentation of scenes and automated prediction of probably fixated objects; collecting a dataset with $>11,000$ gazes containing: raw scenes from head-mounted eye-tracker, corresponding object-mask annotations, estimated fixation regions and probable fixated objects; developing an approach to estimate probability distributions over fixated objects using overlap between `fixation region' (from spatial accuracy of eye-tracking camera) and semantic segmentation from the MaskRCNN

\section{Background}
\label{sec:background}
Dogs generally have less sensitive color perception and visual activity than humans, though they are more sensitive to flicker rates and surpass human visual performance in dim lighting conditions \cite{Byosiere2018}. Previous work has also analyzed how dogs respond to stationary visual stimuli with screen-based eye-tracking \cite{Karl2019}\cite{Somppi2012} by presenting precise stimuli and recording dogs’ eye movements in controlled conditions. These works have shown that dogs tend to direct their visual attention to living creatures in the foreground (vs. the background) – a pattern also observed in chimpanzees and humans \cite{Kano2013}\cite{Törnqvist2020}. Outside of the social domain, other works have shown that dogs are sensitive to physical principles of the world, such as contact causality \cite{Völter2021}. However, most of these works employ screen-based methods in controlled environments and perform visual analysis using low-level saliency e.g. lighting, color etc… Hence it is not currently known how dogs allocate visual attention to objects in their field of view within open-ended natural environments.

Head-mounted eye-tracking is typically used to study visual attention in natural environments. These systems capture the wearer’s first-person view of the world, and record their eye movements; the intrinsics and extrinsics of cameras in the system are used to calibrate eye movements to an approximate gaze location within the user’s field of view. Head-mounted eye-tracking has previously been used to compare cross-species visual behavior \cite{Einhäuser2009}, comparing how cats and humans coordinate eye and head movements. Studies \cite{Huang2022} have also used segmentation models to mask the pupils of small mammals (in closed environments) from eye-tracker images, which is used to synchronize eye-tracking and electroencephalogram recordings. Other works \cite{Machado2011} \cite{Jacob2021} examined the visual attention of monkeys to specific regions of interest in pre-defined digital stimuli. These works observe a subject's visual attention when performing particular tasks e.g. viewing digital stimuli of a monkey face or performing same-different tasks.

Our approach is less-invasive and enables continuous analysis of visual attention in open-ended real world environments (no digital stimuli) using the `semantic' space of objects in the scene; we also analyze visual attention more generally, than conditioned on specific tasks. We collect dogs’ field of view during a walk in a college-city environment along a predetermined route. Our study aims to develop an end-to-end pipeline for semantic classification of fixated objects from visual feedback; we also use this pipeline to categorize the visual information available to dogs and analyze how they allocate their attention within their space. Our statistical analysis had 4 aims:\\
1) To explore the frequency of objects present in the dogs’ field of view to understand the objects available for dogs’ attention in an open-ended environment\\
2) To understand how consistently dogs looked at object classes when they were present in the environment; we evaluate the length of time dogs fixated on a class object relative to the length of time the object was in their field of view\\
3) To analyze differences in visual attention to object classes between individual dogs by considering objects in their view and the object that is attended to\\
4) To compare traditional visual attention models based on saliency with our proposed visual attention model for dogs, based on semantic objects\\
The MaskRCNN is used to generate 2D masks for each target object in the image at the pixel level. Object segmentation is challenging in this data domain, due to the low resolution of head-mounted eye-tracking camera and the limited number of available data points i.e. $~600$ training images for 4 of 11 the dog participants or  $<20\%$ of all data collected. The calibrated eye-tracking data is integrated with MaskRCNN predictions to provide a label predicting the class object the dog was attending to at each fixation. 

\section{Methodology}
\label{sec:method}

\textbf{Ethical Note: }This study was approved by the University’s Institutional Animal Care and Use Committee (IACUC). Study procedures were in accordance with the ASAB/ABS Guidelines for the Use of Animals in Research and complied with the United States Department of Agriculture Animal Welfare Act \& Regulations. All pet dogs were walked by their guardians throughout the session.\\

11 dogs (Female = 7, Mean Age = 6.5 Years) were recruited for participation in this study. Prior to participation, dogs were trained by their guardians to wear commercially available dog goggles in order to acclimate to the eye-tracking goggles following methods in \cite{Pelgrim2022}. Guardians approved dogs to participate when their dogs were comfortable walking outdoors wearing training goggles for at least 10 minutes. Prior acclimatization minimizes visual disturbances and occlusions due to the dogs’ unfamiliarity with the eye-tracking apparatus.

\subsection{Raw Data Collection}
\label{sec:data-collection}
Raw data was collected in $2$ phases: first, an `annotated dataset' was collected, annotated and separated into training/test subsets to fine-tune the MaskRCNN, using fixations from $4$ of $11$ dogs; second, a `non-annotated dataset' was collected using fixations from the remaining $7$ dogs, on which the fine-tuned MaskRCNN made predictions in inferece without a ground-truth. During walks through the environment, dogs wore a head-mounted eye-tracker with two cameras affixed to a dog goggle (Positive Science, Inc.). The infrared `eye camera' recorded the dog’s right eye using an infrared diode. The second `scene camera' recorded the dogs’ first-person view of the scene with field of view of 101.55º horizontally and 73.60º vertically. Video feed from both cameras were digitized at 29.96 fps.

\begin{figure}[t]
  \centering
   \includegraphics[width=0.9\linewidth]{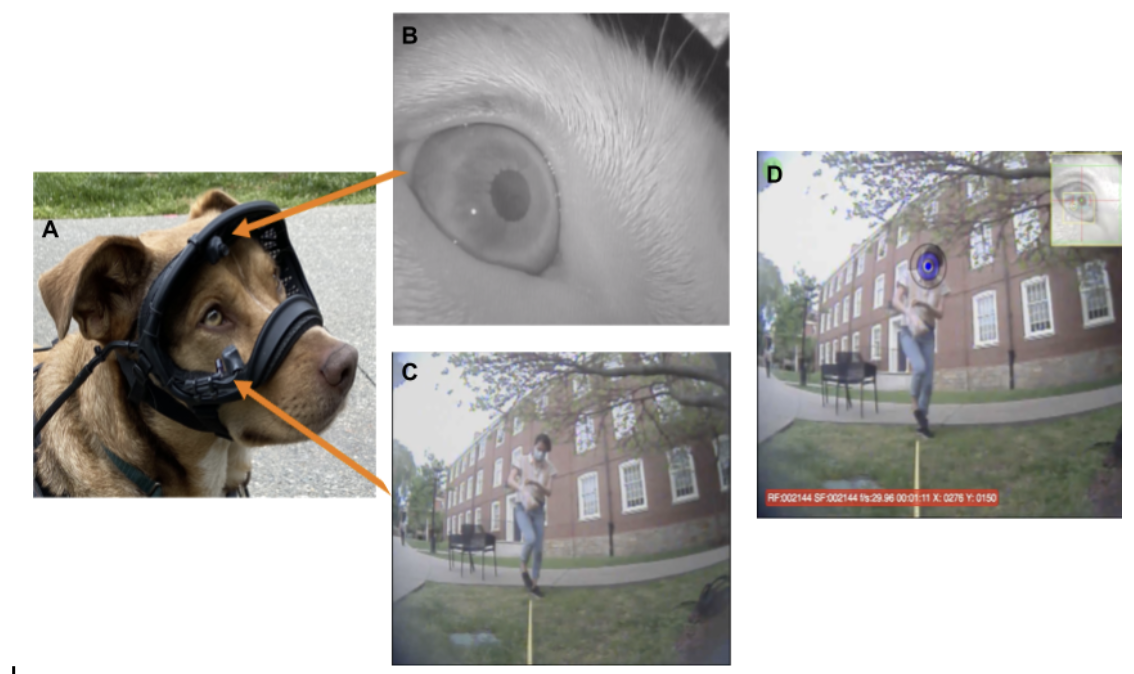}

   \caption{\textbf{A.} A dog wearing eye-tracking goggles with two cameras, \textbf{B.} The dog’s eye recorded from the eye camera, \textbf{C.} The dog’s view recorded by the scene camera. \textbf{D.} The dog’s view with their point of fixation indicated by the blue circle.}
   \label{fig:eye-camera}
\end{figure}

To calibrate the cameras on the eye-tracking apparatus, each dog followed 5 anchor points in space via eye movements alone, as part of a calibration procedure before the walk. The 5 anchor points were spread widely across the dogs’ first-person view, thus requiring a wide range of eye movements. This calibrated the `eye camera's' and `scene camera's' extrinsicts, for grounding the dogs’ visual gaze to a point on the scene camera capturing their field of view, which could be recorded for further offline evaluation \cite{Pelgrim2022}.

Following initial calibration, dogs walked with their guardians along a predetermined 800 meter route which included a variety of scenery e.g. city streets and quiet campus greenspaces. The head-mounted camera was adjusted and the calibration procedure was repeated at any point when the eye-tracker was disturbed or shifted e.g. the dog shook their body or brushed against a wall.

\subsection{Raw Data Pre-processing}
\label{sec:raw-data-prep}
Data recorded from the eye and scene cameras were combined to identify the duration (start and stop in ms) and point of regard (x-y coordinates from the scene camera) for every visual fixation. Fixations were defined as a stable eye positioning lasting $\geq100$ milliseconds.

Previous studies using the eye-tracking system reported spatial accuracies of 2-4º in humans \cite{Franchak2011}\cite{Watalingam2017}, 4º in peacocks \cite{Yorzinski2013} and 3.6º in dogs indoors \cite{Pelgrim2022}. The error between the point of regard extrapolated from the eye-tracking system and the known point of regard (5 anchor points) during calibration was averaged across 20 frames for each dog; the average spatial accuracy for the system across 11 dogs was 5.32º. This is less precise than previous work, likely due to a combination of the outdoor deployment of the system (for the first time) under variable lighting and the low resolution from the scene camera. Each dog's unique radius of error was accounted for when constructing the  fixation region, see \ref{subsec: automatic-identification-of-fixaitons}, to predict objects of fixation with the fine-tuned MaskRCNN segmentations. Target objects generally occupied large enough portions of the frame that the spatial accuracy ($1\%$ of the entire image; approximately the same occupancy as the smallest target object) did not significantly impact results. Finally, instances where dogs were relying on another sense, e.g. sniffing bouts, were filtered out of the raw dataset during the data annotation phase (for the 'annotated dataset') or based on MaskRCNN segmentation prediction (for the 'non-annotated dataset'). Sniffing bouts were defined as frames where fixations occurred but $\leq2$ object masks were annotated/predicted in the scene. Given the diversity of objects in the annotated dataset ($5.03\pm1.14$ objects on average per frame) frames with $\leq2$ objects were $>2\sigma$ from the mean and most likely too close to the camera for dogs to be visually considering the scene.

\subsection{Data Annotation}
\label{sec:data-annotation}
\begin{figure}[t]
  \centering
   \includegraphics[width=1.0\linewidth]{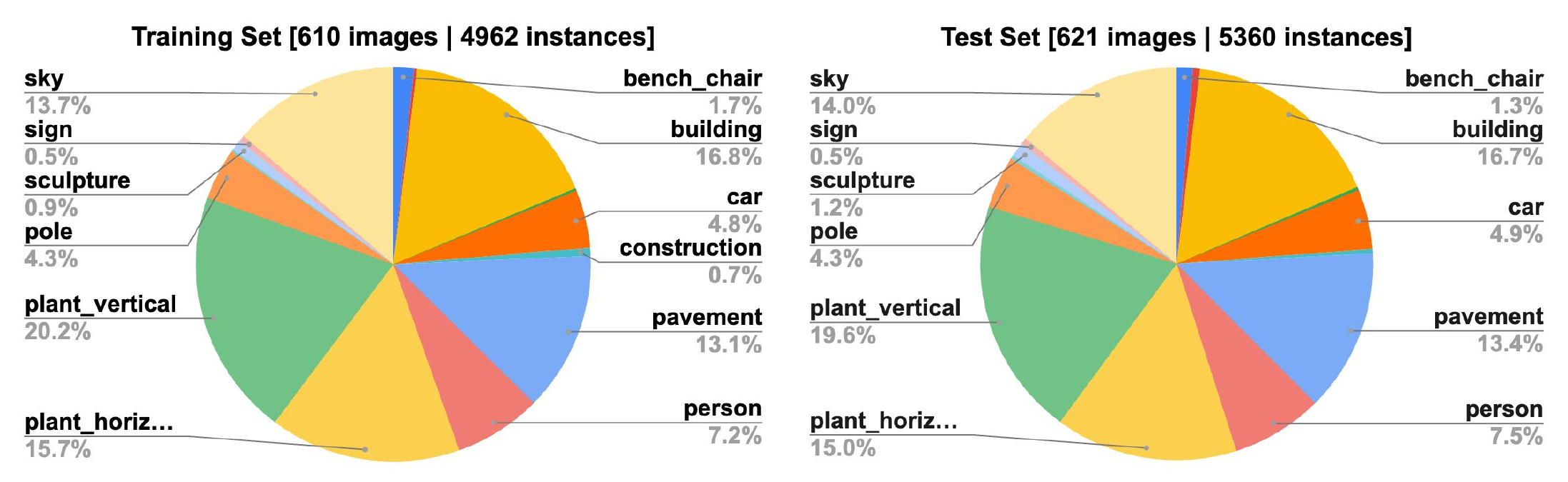}

   \caption{The percentage distribution across object classes in annotated dataset used for training (left) and testing (right) the MaskRCNN. The class distribution between 610-image training set and the 621-image testing set is roughly identical}
   \label{fig:dataset-distribution}
\end{figure}

Before fine-tuning the MaskRCNN via transfer learning, we filtered the `annotated dataset' for frames that demonstrated fixations and removed sniffing bouts as in \ref{sec:raw-data-prep}, resulting in an 'annotated dataset' with 1231 images from $4$ of the $11$ participants. All images were manually annotated with ground-truth pixel-level masks for the $15$ target classes using LabelStudio\footnote{https://labelstud.io/}, an open-source annotation software and split evenly into a training and test subset. This ensured to capture only the object classes of interest in all images, the distribution of which is shown in Figure \ref{fig:dataset-distribution}.

\subsection{Instance Segmentation using MaskRCNN}
\label{sec:instance-seg-mrcnn}
To automate the semantic labelling of objects in the dog’s field of view, we fine-tuned a MaskRCNN for instance segmentation. The model performs both object detection (generating semantic labels to identify objects in the dog field-of-view and their relative sizes) and instance segmentation (generating “masks” identifying the pixels of the image corresponding to different semantic objects). This, in conjunction with the radius of error from the eye-tracking camera, allows us to predict a probability distribution over potential objects the dog is fixated on.

The MaskRCNN uses a ResNet-101 Feature Pyramid Network (FPN) backbone and is pre-trained on MS COCO \cite{lin2015microsoft} dataset. Transfer learning is used for our segmentation task in a novel image domain by fine-tuning the MaskRCNN on 610 images in our training set (of 1,231 images in the annotated dataset) using gradual freezing: for the first 15 epochs, we fine-tune the region proposal network (RPN), final layers (layers $\geq 4$) of the FPN, and the mask generation and object classification heads; for the last 5 epochs, we only continue to fine-tune the mask generation and object classification heads. We empirically observed that gradual freezing yielded optimal results; several previous works \cite{Cetinic2018} \cite{Adiba2019} \cite{Heker2019}, also fine-tuning segmentation models, have shown gradual freezing/un-freezing mitigates catastrophic forgetting and stabilizes learning. Transfer learning enables SOTA performance with limited fine-tuning \cite{Krizhevsky2012}\cite{Pan2010} and generally improves accuracy on smaller novel data distributions that would cause deep neural networks to overfit if trained from scratch \cite{Romero2019}. Previous work \cite{Oquab2014} has demonstrated that CNNs trained on large datasets (e.g. ImageNet) learn useful generic low and mid level representations that can be transferred to new domains with limited data. Using pre-trained weights significantly reduced time to convergence when fine-tuning on our training set.

During 20 epochs of fine-tuning (15 epochs for the first phase and 5 on the second) we used stratified learning rates of $4\times10^{-3}$ and $2\times10^{-3}$ for each phase. Other hyper parameters included: Non-Maximum Suppression (NMS) threshold = 0.85; ROI threshold = 300; 120 batches per epoch; batch size = 5 images. The MaskRCNN's total loss ($L_{MRCNN}$) is adopted from \cite{Kaiming2015}. This loss was also multiplied by a coefficient that scaled inversely with the Mask R-CNN’s prediction confidence and was only applied when the predicted and ground truth class labels match for bounding box pairs with IoU larger than a threshold (IoU$\geq 0.75$). This weighs total MaskRCNN loss more heavily for under-represented classes that otherwise maintained lower prediction confidence and low-quality mask outputs, even if bounding-boxes overlapped significantly, after fine-tuning. The modified loss is shown in Figure \ref{fig:original-vs-new-loss}

 \begin{figure}
 
 \text{original loss:} $L_{MRCNN} = L_{cls} + L_{box} +L_{mask}$\\
    \text{modified loss:} $L_{MRCNN} = (1-C)*(L_{cls} + L_{box} +L_{mask})$\\
 \caption{Our modification to the original Mask-RCNN loss proposed by \cite{Kaiming2015}. We add a coefficient to the loss that scales inversely with the class confidence $C$, when the bounding box and class predictions match the ground truth}
 \label{fig:original-vs-new-loss}
 \end{figure}

After transfer learning, the MaskRCNN is evaluated on the 621 unseen test set images sampled from the 1,231 image annotated dataset. The fine-tuned MaskRCNN's performance is analysed in \ref{subsec:mrcnn_eval} across $5$ evaluation metrics. The MaskRCNN can now make reliable segmentations of scenes in the novel image domain. Next, a system to estimate fixated objects was developed (in \ref{subsec: automatic-identification-of-fixaitons}) by considering the overlap between the probable ``fixation region'' produced by the eye-tracking camera's radius of error around the estimated fixation point and the fine-tuned MaskRCNN's predicted segmentations.

\subsection{Automatic Identification of Fixated Objects}
\label{subsec: automatic-identification-of-fixaitons}
The fine-tuned Mask-RCNN and the eye-tracking camera data captured for the non-annotated dataset were used to predict a probability distribution over potential fixated objects for all fixations generated by the remaining of $7$ of $11$ participants. These datapoints served as an `in the wild' application of the fine-tuned MaskRCNN for fixation prediction without ground-truth data

\begin{figure}[t]
  \centering
   \includegraphics[width=1.0\linewidth]{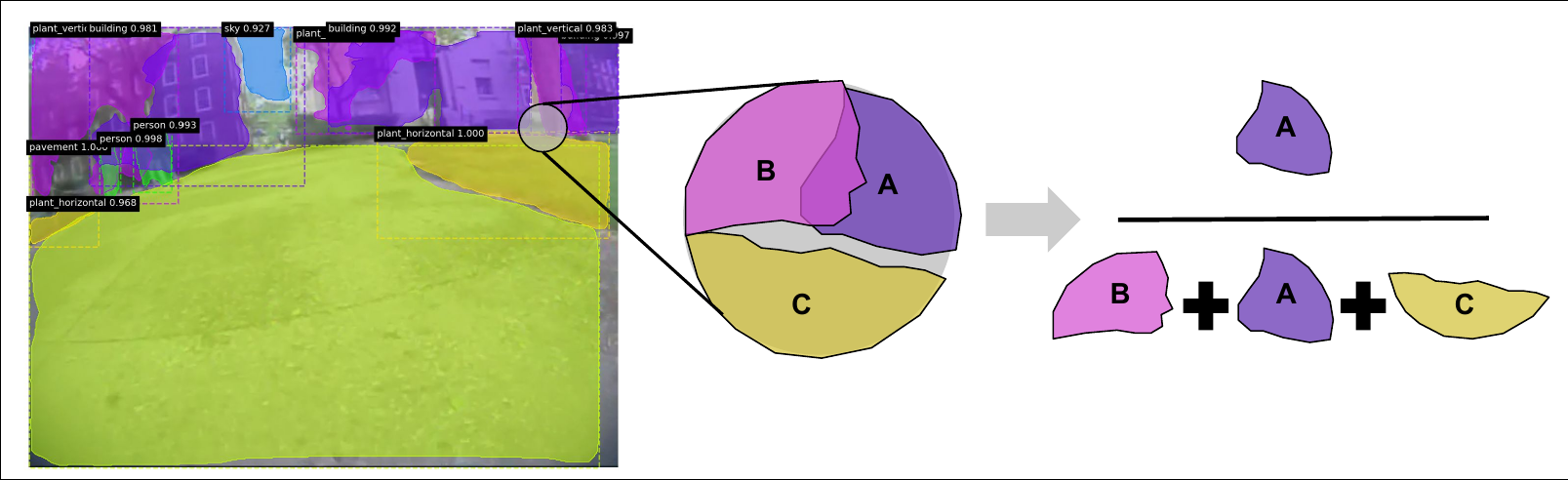}

   \caption{Procedure to estimate the probability distribution over fixated objects. (Left) Given the “fixation region” due to the radius of error, (Center) we consider portions of predicted object masks that intersect the fixation region. (Right) We estimate the probability of fixating to a given class (class ‘A’)  by normalizing total number of pixels in the intersection belonging to the given class (no. of pixels for ‘A’) over total number of pixels belonging to all classes in the intersection (no. of pixels for ‘A’, ‘B’ and ‘C’)
}
   \label{fig:fixation-prediction}
\end{figure}

Using the unique spatial accuracy from the eye-mounted camera for each dog, we extended a `radius of error' around the estimated fixation pixel i.e. we assumed the true fixation point lay within this circular `fixation region' with a radius equal to the `radius of error'. To estimate the probability that a particular fixation attended to object class $A$, we considered the subset of predicted masks overlapping the fixation region and counted the total number of pixels in the intersection belonging to predicted masks for class $A$; this pixel count (for class $A$) was normalized by the total count of pixels for \textit{all} object classes (number of pixels belonging to predicted masks for classes $A,B,C,...$) in the intersection. Figure \ref{fig:fixation-prediction} visually highlights the process. Repeating the process, to estimate the probability that each of the $15$ object classes of interest is being attended to, generates a probability distribution over possible fixated objects

The same approach is used to assess how similar the probability distribution over fixated objects is using masks predicted by the MaskRCNN compared to using ground-truth masks, for the 610 image test set. The chi-square distance between the probability distributions (using predicted masks and ground-truth masks) was computed for each image in the test set. This was used to determine the similarity of the MaskRCNN’s predicted masks to ground-truth masks within the “fixation region”, thereby assessing if the MaskRCNN is a good estimator for the probability distribution of possible fixated objects.  

\section{Results \& Discussion}
\label{sec:results}
Across 11 dogs, 11,698 fixations were recorded spanning 102,868 frames. The fine-tuned Mask-RCNN eliminates the need for any future annotations on the non-annotated dataset (from 7 of the 11 dogs) reducing total time required to segment objects and identify possible fixated objects in all frames by $>200\times$. To measure generalization to the new data distribution, the fine-tuned MaskRCNN was evaluated on a test set of 621 images captured by 4 of 11 participants, across 5 metrics: class-based accuracy, intersection-over-union (IoU) of masks, mask-based recall, mask-based precision and distribution of chi-square distances. The evaluation of fine-tuned MaskRCNN is presented in section \ref{subsec:mrcnn_eval} below. Quantitative analysis on visual behavior was then performed across the broader dataset of 11 dogs (11,698 fixations) after eliminating sniffing bouts and any fixation frames with null predictions for fixation objects – leaving 10,296 fixations for quantitative analysis. The evaluation on visual behaviour is presented in section \ref{subsec:analyze-visual} below.

\subsection{MaskRCNN Evaluation}
\label{subsec:mrcnn_eval}

\begin{figure}[t]
  \centering
   \includegraphics[width=0.9\linewidth]{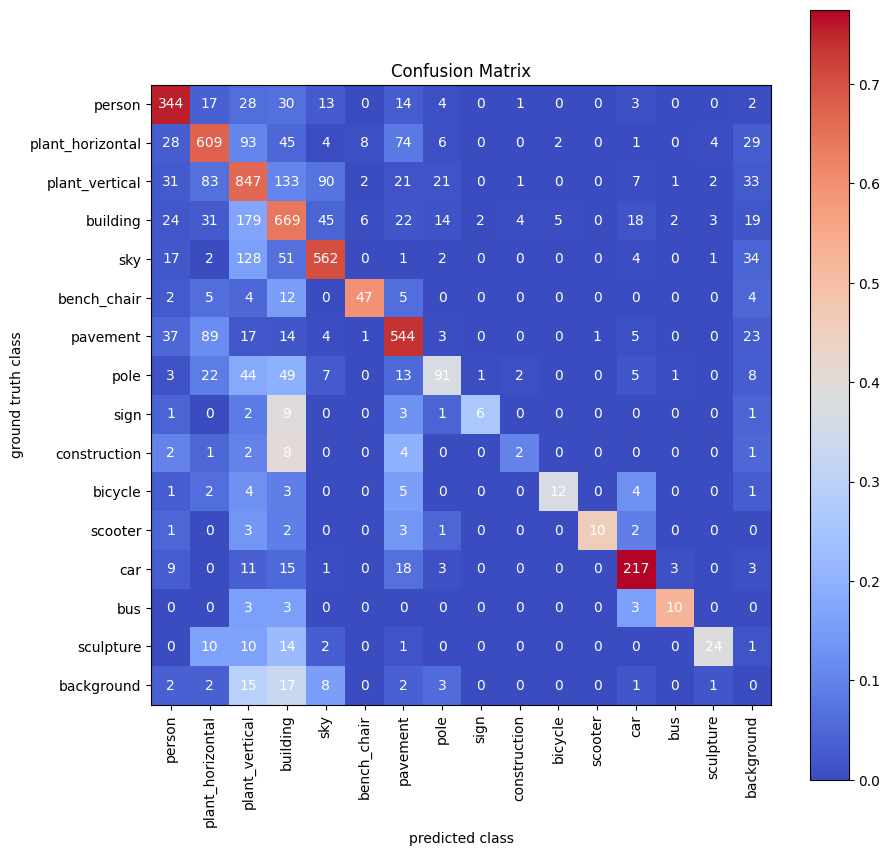}
    \caption{Confusion matrix for predicted (x-axis) v.s. ground-truth (y-axis) class. The color bar indicates the distribution predicted classes for ground-truth instances of a certain class}
   \label{fig:conf-matrix}
\end{figure}
\begin{figure}[t]
  \centering
   \includegraphics[width=0.8\linewidth]{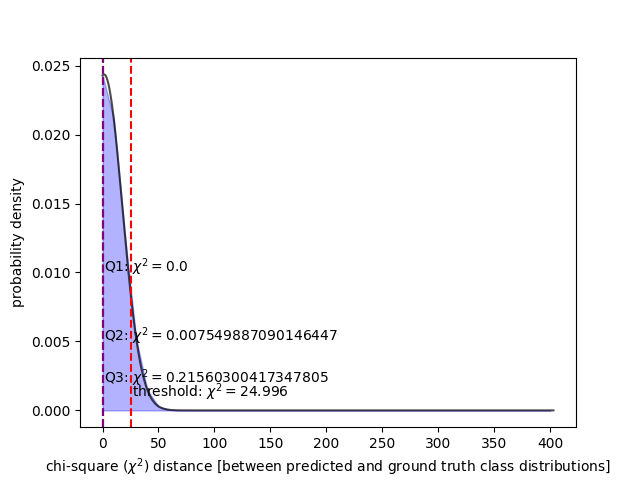}
   \includegraphics[width=0.8\linewidth]{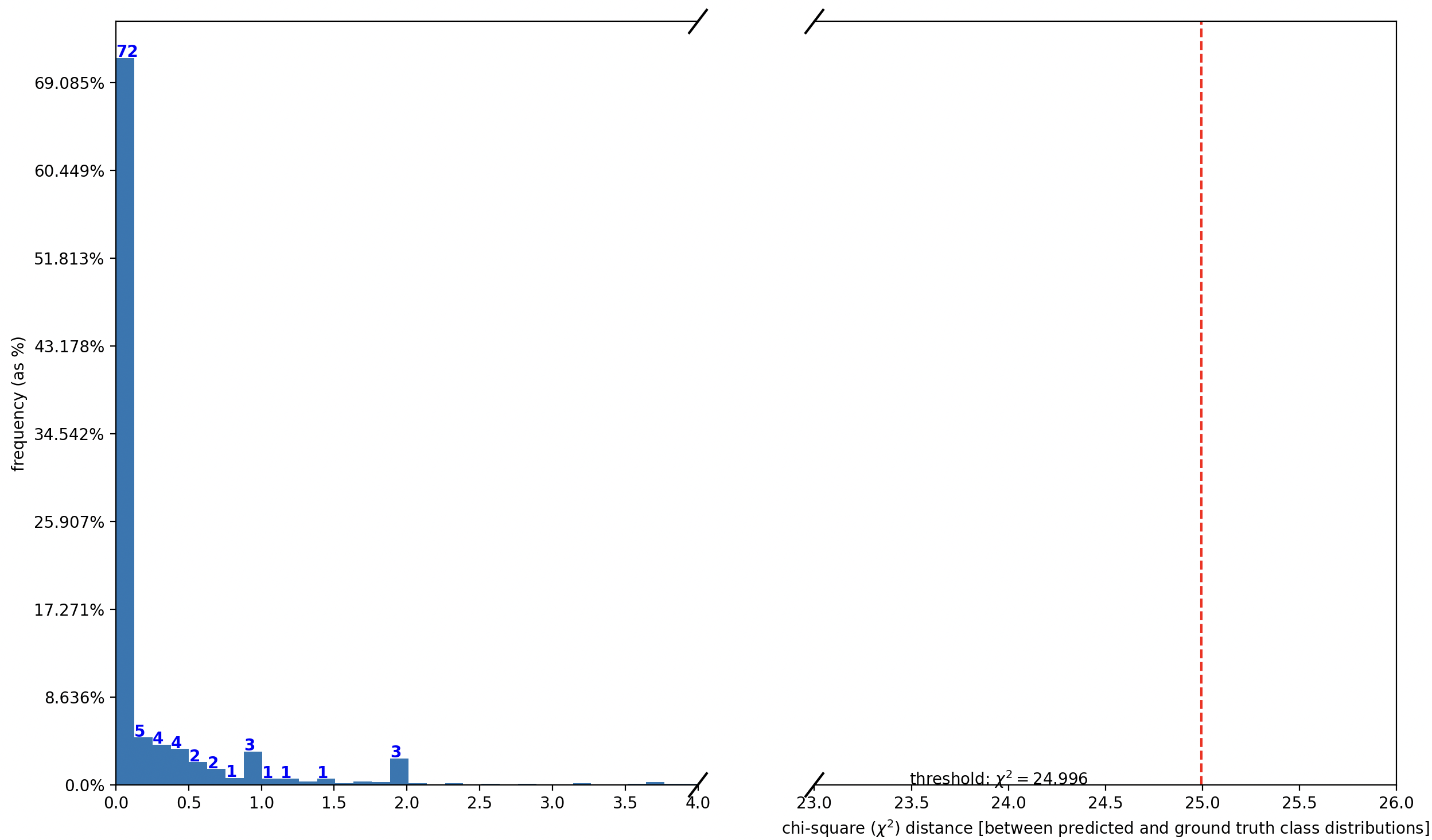}
    \caption{Top: chi-square distances for the test-set fit to a normal distribution highlighting the Q1, Q2, Q3 chi-square distance; Bottom: chi-square distance distribution for the range $0-25$ i.e. upto the critical value for DOF$=15$}
   \label{fig:chi-square-dist}
\end{figure}
\begin{figure}[t]
  \centering
   \includegraphics[width=0.95\linewidth]{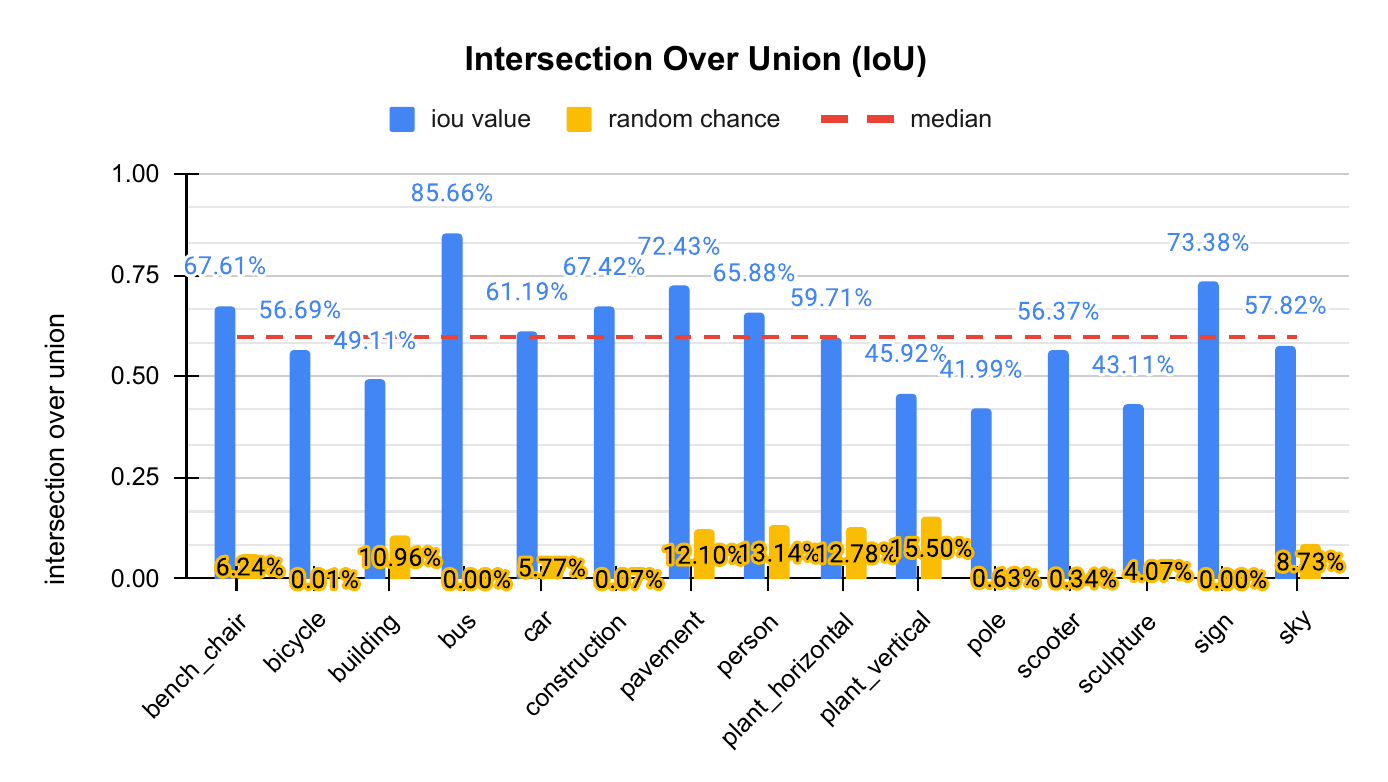}
    \includegraphics[width=0.95\linewidth]{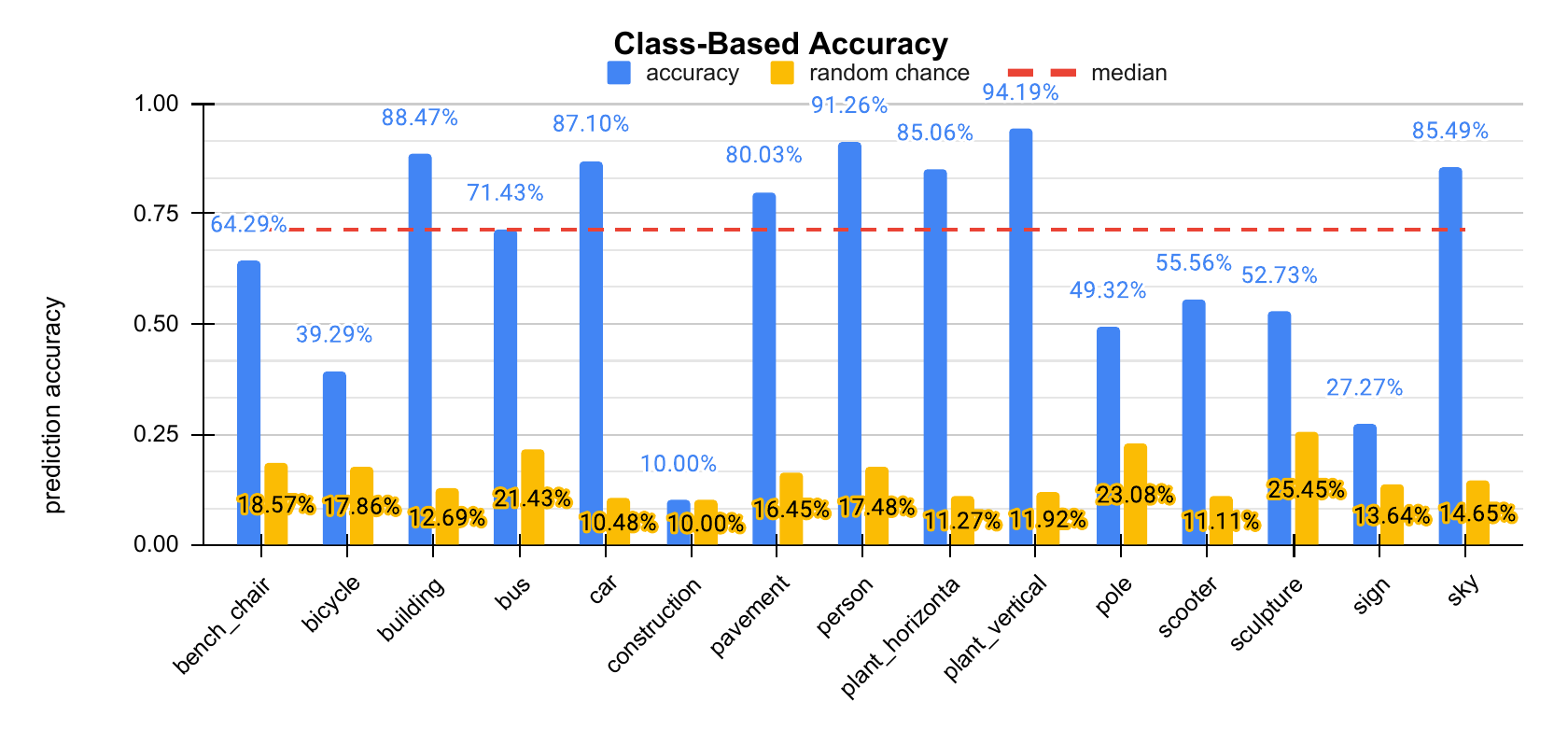}
\caption{Mean IoU and accuracy per class averaged across all instances of the class in the test set}
   \label{fig:metrics}
\end{figure}

Figure \ref{fig:metrics} measures the class-based accuracy for the ‘class prediction’ task: comparing counts of ground-truth classes with the counts of predicted classes for each image, without accounting for spatial overlap of pixel-level masks. The metric measures whether class predictions of the MaskRCNN match ground-truth, even if predicted masks may not. The confusion matrix in Figure \ref{fig:conf-matrix}, instead, measures misclassifications for the ‘mask prediction’ task: every ground-truth mask (and its associated class label) is paired with the predicted mask (and its associated class label) having maximum IoU and vice-versa, to account for false positives. The metric measures whether both the MaskRCNN class predictions and generated masks match ground-truth. This is a much stricter metric for quantitative performance since classes that dominate the scene background (sky, plant vertical, pavement or background) have generally large IoU with most object masks due to their mask sizes; so even slight misalignments between predicted and target masks for smaller objects can cause maximum IoU to default to one of these background classes. The mask-based accuracy, true positive rate and precision metrics are all derived from the confusion matrix for the stricter ‘mask prediction task’.

\textbf{Average coverage of predicted masks over the frame for all identified objects}. The fine-tuned MaskRCNN generates masks that sometimes occupy a subset of the pixels occupied by ground-truth masks, though this does not significantly impact the estimated probability distribution over potential fixation objects. Masks generated by the MaskRCNN covered on average $84\%$ of the image ($9\%$ less) compared to $93\%$ of the image covered on average by ground-truth masks. The ‘pavement’, ‘construction’ and ‘sign’ classes most significantly contributed to this difference,  with predicted masks under-predicting mask sizes by $-3.36\%, -2.47\% \text{and} -1.98\%$ respectively. Differences in assigned class probabilities within the fixation region are much smaller (on average $-2.92\%$) with the ‘construction’ ($-14.29\%$) class being the most under-assigned class within the fixation region and ‘bicycle’ ($+7.43\%$) being the most over-assigned class within the fixation region. The average ($-2.92\%$) difference suggests the MaskRCNN may be slightly more likely to generate false negatives than false positives within the fixation region.

\textbf{Fine-tuned model demonstrates statistical significance in matching ground-truth probability distributions} for object classes present in the “fixated region” across all 621 test images. The 90th percentile of chi-square distance between predicted and ground-truth masks (within the fixation region) lies $\leq1.0$ (see Figure \ref{fig:chi-square-dist}) compared to 90th percentile chi-square distance of $3.5$ for random chance. For a significance of $\alpha=0.05$ and $15$ degrees of freedom (15 object classes + background - 1), the chi-square goodness of fit test allows us to accept the null-hypothesis that predicted class distributions match the ground-truth distribution within the region of fixation with confidence $p < .05$. The critical chi-square value (24.996) for the degree of freedom imposed is above the chi-square distance for $> 99\%$ of the $621$ test set images. Hence, the fine-tuned MaskRCNN can reliably serve as an estimator of fixed objects within regions of fixation.

\textbf{Intersection Over Union (IoU) of predicted masks}. The number of pixels overlapping the ground-truth mask, normalized by the number of pixels in the union of both predicted and ground-truth masks. For each class, we only consider the set of ground-truth and predicted masks having the class label and report the ground-truth/predicted mask pairing that maximizes IoU. The MaskRCNN achieves IoU substantially greater than random-chance – indicating generated masks are semantically relevant and largely match ground-truth masks. The median IoU across all classes is $60\%$ ($10\times$ higher than chance) with a maximum IoU of $85\%$ for 'bus' and a minimum IoU of $42\%$ for `pole'. Comparatively, the maximum IoU with chance is $8\%$ for the `sky' class; the smallest improvement of the fine-tuned MaskRCNN over random chance is by $3\times$ for the `plant vertical' class.

\textbf{Qualitative results} in supplementary material also show the fine-tuned MaskRCNN’s capability to segment multiple foreground and background objects in complex scenes with $>10$ object instances. The MaskRCNN is capable of relatively high quality segmentation despite low image resolution, limited color contrast (grays or monotone tones) and the fisheye lens distortion present in the images.

\textbf{Mask-RCNN’s mask-based accuracy, recall and precision} for the mask-prediction task using the confusion matrix. The class-based accuracy is also compared to determine quality of semantic class prediction – without considering mask generation. The MaskRCNN achieves a mask-based accuracy of $66\%$ ($6\%$ for chance), a median TPR of $68\%$ ($2\%$ for chance), a median precision of $56\%$ ($6\%$ for chance) as well as a median class-based accuracy of $74\%$ ($15\%$ for chance). The outlier class “construction” (i.e. construction sites, scaffolding or regions within traffic drums and/or caution tape) has the lowest class-based accuracy, TPR and precision ($10\%, 20\% \text{and} 10\%$ respectively) which is close to chance and likely due to the object being heavily underrepresented with poor semantic definition in the training set i.e. “construction” incorporates broad and inconsistent visual features making classification challenging. The “background” class achieved $0\%$ precision and recall,  with substantially more false negatives attributed to missed identification of “sky” and “plant vertical” classes.  The object classes with highest class-based accuracy and mask-based precision/recall are “car”, “person” and “pavement” that perform at least $7-10\%$ points higher than the median across classes.

Overall, the class-based accuracy shows us that the fine-tuned MaskRCNN captures nearly all objects of interest in the frame. The chi-square distance distribution and goodness of fit test demonstrate strong overlap with ground-truth distributions within the fixation region. The IoU and coverage metrics show substantial coverage and overlap much greater than chance. The precision and recall metrics show that the most significant mis-classifications occur for under-represented objects (like “sculpture”, “construction” or “sign”) and the largest percentage of false positive predictions are dominant background classes (like “plant vertical” or “building”) occurring with very high frequency in the training dataset. Hence, the MaskRCNN for the most part, matches ground-truth masks and is a good estimator of fixed objects within regions of fixation.

\subsection{Analyzing Visual Behaviour}
\label{subsec:analyze-visual}

\textbf{Objects in Field of View} Our first aim was to categorize the frequency of objects present in dogs’ visual environment. This is necessary to understand how dogs direct their visual attention within the environment. We conducted a 15 (object classes) by 11 (dog identity) ANOVA test to explore the impact of class and dog identity on proportion of time different objects appeard in-frame. The test yielded significant effect of item identity, $F(14, 140) = 172.07, p < .001, \eta_{p}^{2} = 0.95$, confirming that certain objects were more common in dogs’ view, and therefore more available fixation targets. Plants (horizontal and vertical), pavement, buildings, and sky were present for the majority of fixations. People, cars, and poles were available fixation targets a moderate amount of time, $10\% < x < 50 \%$, of fixations. Sculptures, benches/chairs, bicycles, signs, construction equipment, buses, and scooters were rarely in dogs’ view. Table \ref{tab:quantitative-results} summarizes the proportion of time each object category was in view. An average of $5.03\pm1.14$ objects were in view across all 11 participants, with a minimum of 3 and a maximum of 12 objects in view. The ANOVA test yielded no effect on dog identity, $F(10, 140) = .879, p = .55, \eta_{p}^{2} = 0.06$, meaning no individual dog encountered substantially different object classes for substantially different proportions of time. We also consider size of objects in dogs’ field of view as the percentage of total pixels in the image occupied by the mask for an individual instance of an object. Sculptures and benches/chairs consistently appeared small in dogs’ views, occupying $1\pm4\%$ and $1\pm1\%$ of the frame, respectively. Pavements and plant horizontal occupied large portions of dogs’ view varied greatly in size e.g. $34\pm9\%$ and $18\pm6\%$ of the frame, for “pavements” and “plant horizontal” respectively. 

\textbf{Allocation of Visual Attention to Objects} Our second aim was to explore how dogs’ allocated visual attention to objects in their view. If dogs look randomly within their view, we would expect uniform fixations across all classes; otherwise dogs may be actively directing their visual attention to selectively attend to certain objects more than others. Following the approach of establishing a fixation region, described in \ref{subsec: automatic-identification-of-fixaitons}, the objects dogs fixated on was determined using the predicted fixation probabilities –  weighted by the percentage of the fixated region each class occupied (see \ref{tab:quantitative-results})

Table \ref{tab:quantitative-results} clearly shows that dogs do not uniformly attend across objects in their view. For each of the 10,296 fixations we used the set of fixation object classes predicted by the MaskRCNN and the region of fixation extrapolated from the eye-tracking camera. Because the potential fixation objects in the region of fixation were represented as a probability distribution over classes, a given fixation could have multiple targets. Hence we conducted a logistic regression with a firth’s correction to explore the relationship between probably objects of fixation as a function of the object class, dog identity, and an interaction between dog and object class. We found a significant effect of object class, $\chi^{2}(14) = 1348.97 , p < .001$ suggesting that dogs do not passively observe their field of view but rather actively directing their attention to certain objects. For example, dogs looked at people on average $15.7\%$ of the time they appeared in-frame, whilst they looked at benches/chairs only $1.2\%$ of the time they appeared in-frame.  Post-hoc pairwise comparisons exploring class differences with a Tukey correction revealed that dogs also looked significantly more to people than other non-social classes like sculptures, $t(51368) = 5.137, p < .001$, and the sky, $t(51368) = 16.346, p < .001$. Dogs also frequently looked to vertical plants (above their body height) $26.9\%$ of the time they were in view. This was significantly more than they looked at people, $t(51368) = -17.154, p < .001$. Surprisingly, though ‘sky’ was nearly ubiquitous in dog’s view – present for approximately $83.8\%$ of fixations – dogs looked to the sky only $7\%$ of the time it was in their view. 

There was significant interaction between dog identity and class on fixation behaviors, $\chi^{2}(100) = 3997.19 , p < .001$, suggests that some dogs fixated more to certain classes than others. For example, one dog (Dog A) looked at people $46\%$ of the time they were in view, whilst another (Dog B) looked at people $4\%$ of the time they were in view. Post-hoc pairwise comparisons using a turkey correction found Dog A looked to people significantly more than other object classes (e.g. buildings, $t(51368) = -5.96, p < .001$, horizontal plants, $t(51368) = 10.843, p < .001$, and sculptures, $t(51368) = 4.60, p < .001)$ whilst Dog B did not look significantly more at people than any other class ($p < .05$) 

We also explored the relationship between dogs’ likelihood to look at objects as a function of their size in the scene.  If dogs are selective in their visual behavior, we could expect visual behavior that occasionally attends to  items that are comparatively smaller in the frame, either due to the object’s relative size or depth. A limitation of our apparatus is the lack of depth data from eye-tracking cameras, which ambiguates object size with proximity.  For $56.75\%$ of fixations, the largest object in-frame was part of the fixation objects predicted by the MaskRCNN's segmentation overlap with the region of fixation. When fixated, the largest object took up an average of $46\pm14\%$ of the dogs’ field of view. A Spearman correlation, found significant positive correlation between average size of object class and time fixated on that class (relative to the time in view) $R = .737, p < .001$. However, a Spearman correlation also found no significant correlation between the average size of object class and the proportion of the fixation region the object occupied when it was fixated upon $R = .147,  p < .001$  i.e. larger objects were fixated for longer but did not necessarily occupy larger portion of the dogs fixation region. However conclusions here should be limited due to the ambiguation of true object size and depth by scene cameras.

\subsection{Evaluating Image Saliency Models}
\label{subsec:image-saliency}

Traditional hypotheses assert that dogs direct their visual attention based on bottom-up processes driven by low-level image features, with preferential attendance to regions of high image saliency – as opposed to directed attention to semantic classes. We perform image saliency analyses using Saliency Toolbox Version 2.3 \cite{Walther2006} in order to understand if a combination of low-level image features (i.e. luminance, color intensity and opponency, orientations) as a saliency map is a reliable estimator of dogs' visual behavior patterns. 

Since dogs have less adept color vision than humans, we first perform image saliency analysis in greyscale and then in full color.  We evaluated the saliency model as an estimator for dogs visual fixation, to determine if their attention was truly driven primarily by low level image features against our fixation predictions based on semantic labels produced by the MaskRCNN; we evaluate the Area Under the Curve (AUC) of the Receiver Operating Characteristics (ROC) for the saliency based model \cite{salMetrics_Bylinskii}.  This compares the accuracy, using true and false positive rates, of the saliency model, relative to the ground-truth fixation data. We used AUC Judd, a variation of AUC where accuracy is increased where saliency map values are above threshold at the fixated region (termed true positives), and penalized by instances where saliency map values are above threshold at unfixed locations (termed false positives) \cite{mit-saliency-benchmark}. We use the definition of fixation region as explained in subsection \ref{subsec: automatic-identification-of-fixaitons}, and compare the average saliency value of that predicted fixation region to the thresholds. 

AUC-Judd was 0.67 when using full-color images and 0.66 for greyscale images, suggesting that the saliency model is a poor predictor of dogs’ visual behaviors, or that it is unlikely dogs’ fixations are driven solely by low-level image saliency features. The ROC curve as well as further details on AUC calculations can be found in the supplementary material

\begin{figure}[t]
  \centering
   \includegraphics[width=0.9\linewidth]{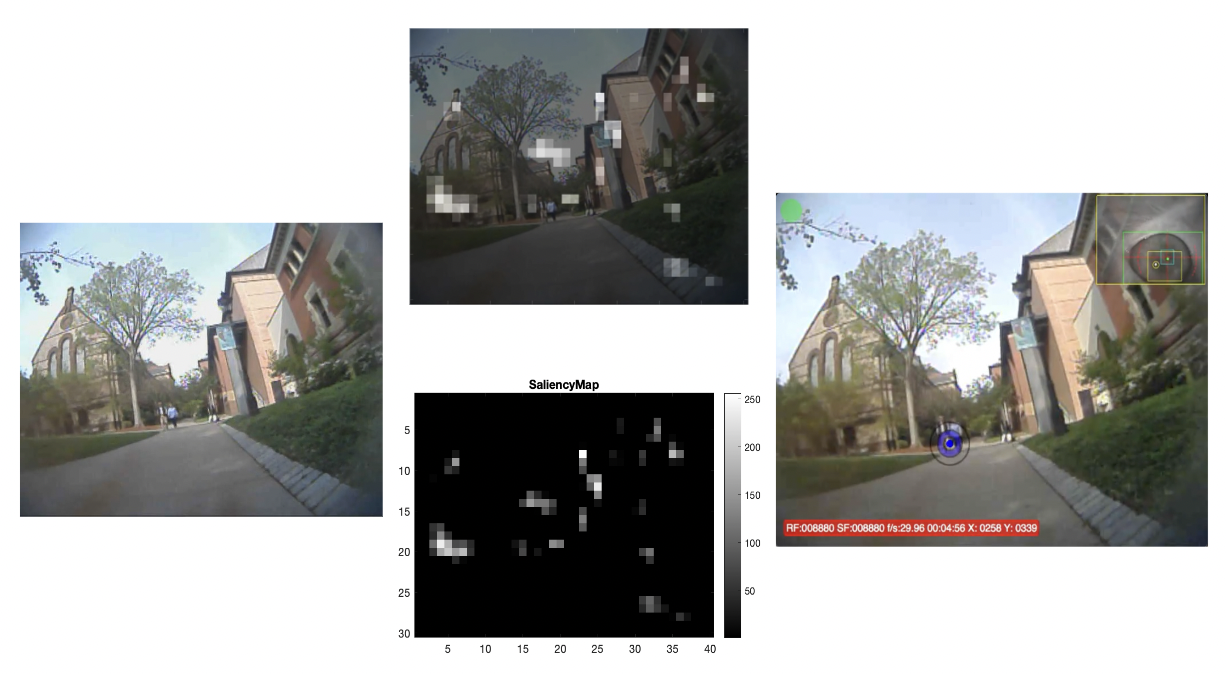}
    \caption{Left: image from the dog’s POV, fed into the Saliency Toolboc; Middle Bottom: saliency map; Middle Top: saliency map overlaid on the dog’s POV; Right: dogs’ fixation (blue circle) which can be examined for saliency value}
   \label{fig:saliency-model}
\end{figure}

\section{Conclusion}
\label{sec:conclusion}
Our study explores the visual behavior patterns of dogs in open-ended natural environments. We collect a dataset of visual feedback from first-person perspective of dogs using a head-mounted eye-tracking camera and perform transfer learning on a pre-trained MaskRCNN on this novel image and class distributions. The fine-tuned MaskRCNN, with the eye-tracking apparatus, enabled significant speed-up to analyses and also acts as a model to predict a probability distribution over potential semantic fixation object that is shown to characterize dogs’ visual behavior better that an image saliency model. Despite training on a limited training set ($< 20\%$ of total data points) the fine-tuned MaskRCNN achieves strong quantitative performance, matches ground-truth masks and is a good estimator of fixed objects within regions of fixation.  Qualitative results also demonstrate that the MaskRCNN is capable of robust segmentation despite image distortions and low resolution. 

Our statistical analysis found that dogs’ visual attention is more selective to certain object classes (people, cars,  plants, construction and pavement) than to other objects. Further, we find suggestions of visual neophilia in dogs, driving the tendency to attend more to unusual and novel objects, such as construction equipment. Finally, we found that, on average, individual dogs were exposed to a similar distribution of objects in their field of view and had similar visual behavior patterns. 

This study was the first to record how dogs visually observe their environment in natural open-ended settings. Whilst we focus on visual attention of dogs, the method can be extended to other species that can acclimatize to the head-mounted eye-tracker \cite{Yorzinski2013}, since the MaskRCNN operates modularly and independently to the eye-camera on the tracker. Future research can expand the set of object classes categorized in this work. One limitation was the spatial accuracy of the eye-tracking system given variable outdoor lighting conditions and lack of depth feedback from the scene camera – to disambiguate foreground background objects. Improving camera resolution can enable future work to identify whether, for instance, dogs look at faces of people rather than hands or feet of people. Understanding visual behaviors patterns of dogs, particularly when performing mobile tasks, would provide insight into designing more social/natural quadruped robots or improving human-dog cooperation. Additionally, analysis on how dogs visually perceive their environment could support unique attention models that are inspired by dogs’ visual cognition and could be utilized for visual decision making or embedding statistical visual preferences into autonomous systems. 

\section{Acknowledgements}
\label{sec:acknowledgements}
Thank you to our participants and guardians. Thank you to our funding sources: NIH (T32MH115895), ONR (N00014-19-1-2029), NSF (IIS-1912280 and EAR-1925481), DARPA (D19AC00015), NIH/NINDS (R21 NS 112743), and ANITI (ANR-19-PI3A-0004). Computing hardware supported by NIH Office of Director (S10OD025181) and Center for Computation and Visualization (CCV). 

{\small
\bibliographystyle{ieee_fullname}
\bibliography{egbib}
}

\newpage
\section{Appendix}
\label{appendix}

\subsection{Appendix A: Qualitative Performance of MaskRCNN}
\label{appendix_a}

\begin{figure}[ht]
  \centering
   \includegraphics[width=1.0\linewidth]{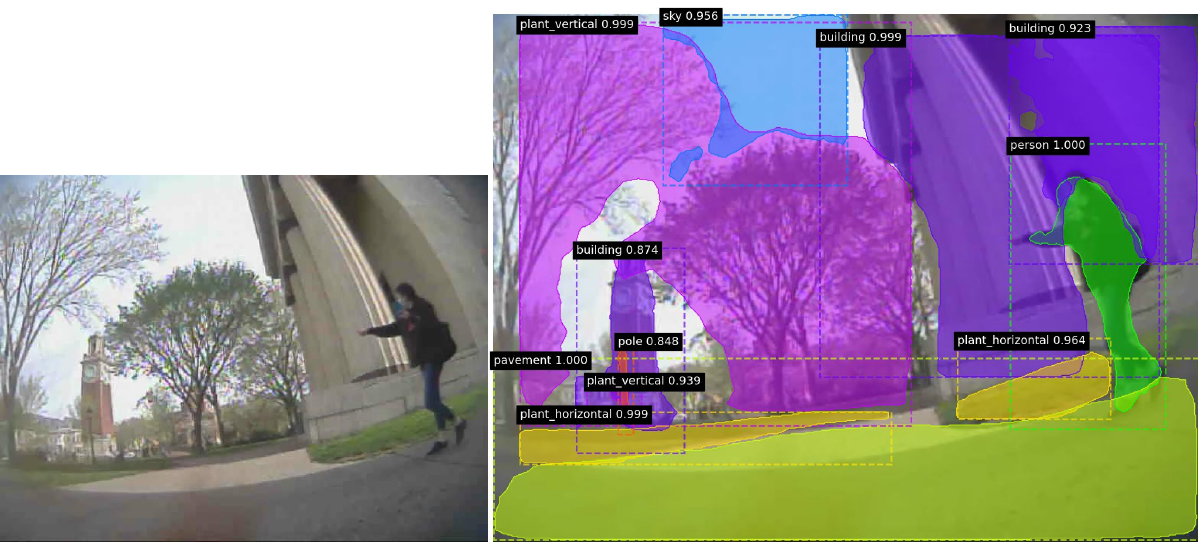}
   \includegraphics[width=1.0\linewidth]{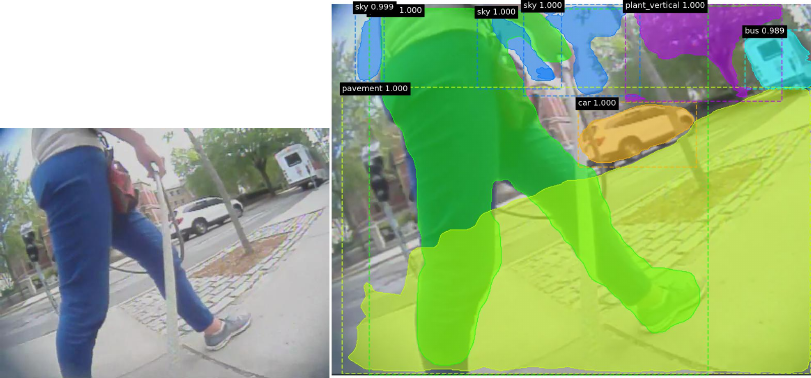}
   \includegraphics[width=1.0\linewidth]{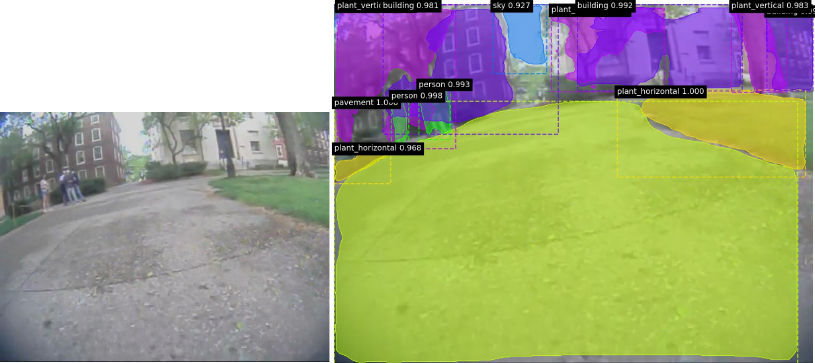}
   \includegraphics[width=1.0\linewidth]{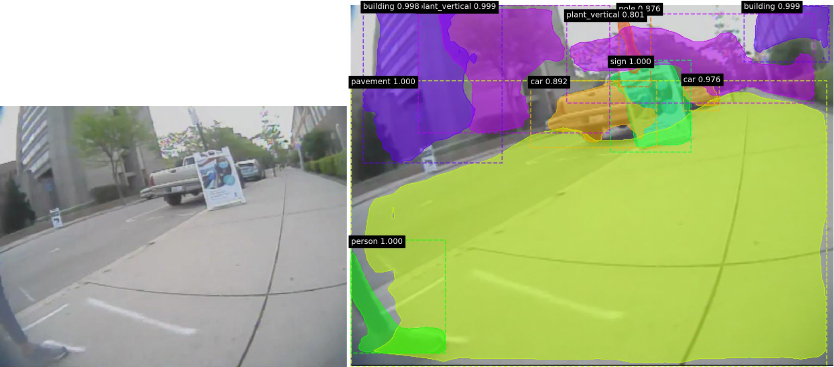}
    \caption{Qualitative examples of instance segmentation by MaskRCNN on images from the test set. Left: original image captured by head-mounted camera; Right: fine-tuned MaskRCNN’s segmentation masks}
   \label{fig:qualitative-examples}
\end{figure}

\newpage

\subsection{Appendix B: Analyzing Qualitative Null Fixations}
\label{appendix_b}

When generating fixation predictions for all 11 participants in the student, ~150 fixation frames ( $< 1.5\%$) of the 11,689 fixations recorded yielded a null fixation prediction. Some examples of these null fixations are shown in Figure 8 below, where the red circle indicates the dog’s region of fixation for the specific frame. These null fixations were generally caused due to one of four misalignments:
\begin{itemize}
    \item the generated segmentation mask for an object did not cover the entire object, and the region of fixation lay on the section of the object that was not segmented
    \item due to low image resolution and fish-eye distortion of the camera, the region of fixation appears blurry and target object is unclear from the eye-tracking scene camera’s point of view
    \item the region of fixation lies on an part of the image with an ambiguous class or a class that was not annotated as part of the 15 target objects considered in the study e.g. top of a fence, the stone pillar to an archway or the dog’s leash
    \item the region of fixation lies on a target object (typically a portion of the ‘sky’) that is particularly small and non-contiguous in the frame, therefore not segmented completely
\end{itemize}

\begin{figure}[h]
  \centering
   \includegraphics[width=0.8\linewidth]{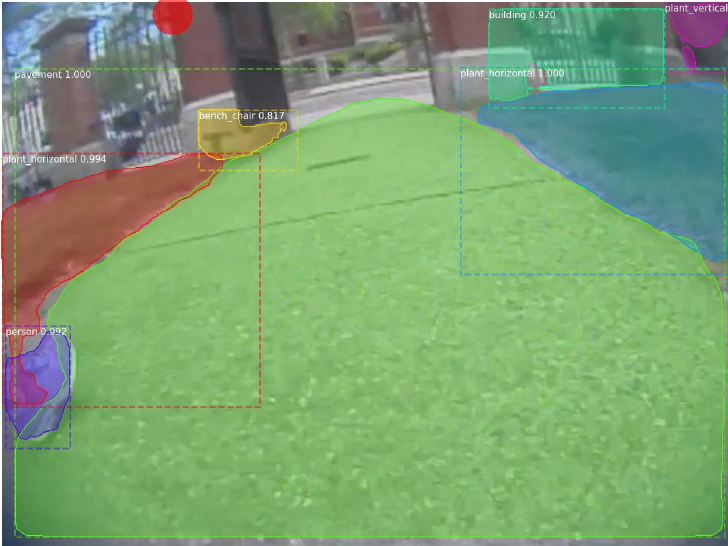}
   \includegraphics[width=0.8\linewidth]{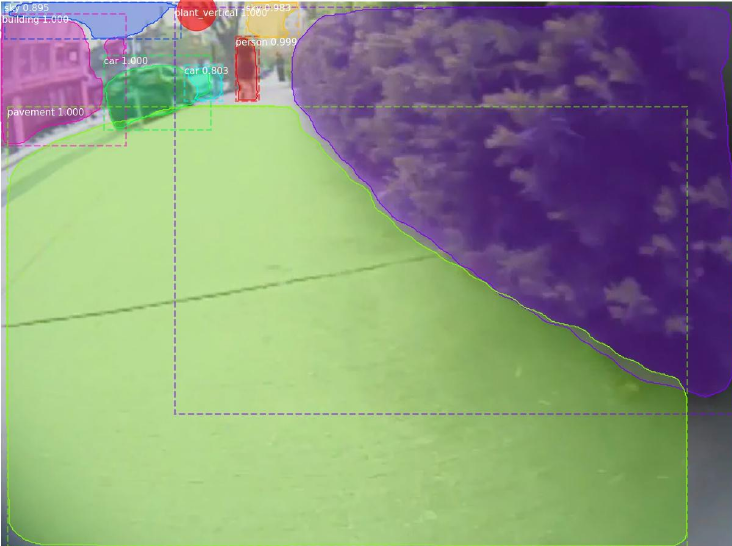}
   \includegraphics[width=0.8\linewidth]{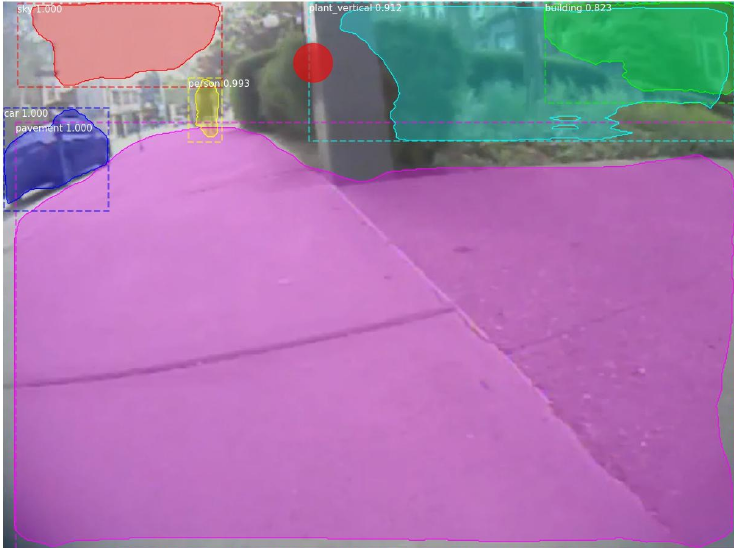}
 \caption{Qualitative examples of null fixations produced by the fine-tuned MaskRCNN when attempting to automatically predict a semantic label for the dog’s visual attention. The red circle represents the possible region of fixation (accounting for calibration error) from the predicted fixation point that does not intersect with any predicted masks.}
   \label{fig:null-fixations}
\end{figure}

\newpage

\subsection{Appendix C: ROC\& AUC comparing image saliency model with semantic segmentation model for Visual Gaze prediction}
\label{appendix_c}

\begin{figure}[h]
  \centering
   \includegraphics[width=1.0\linewidth]{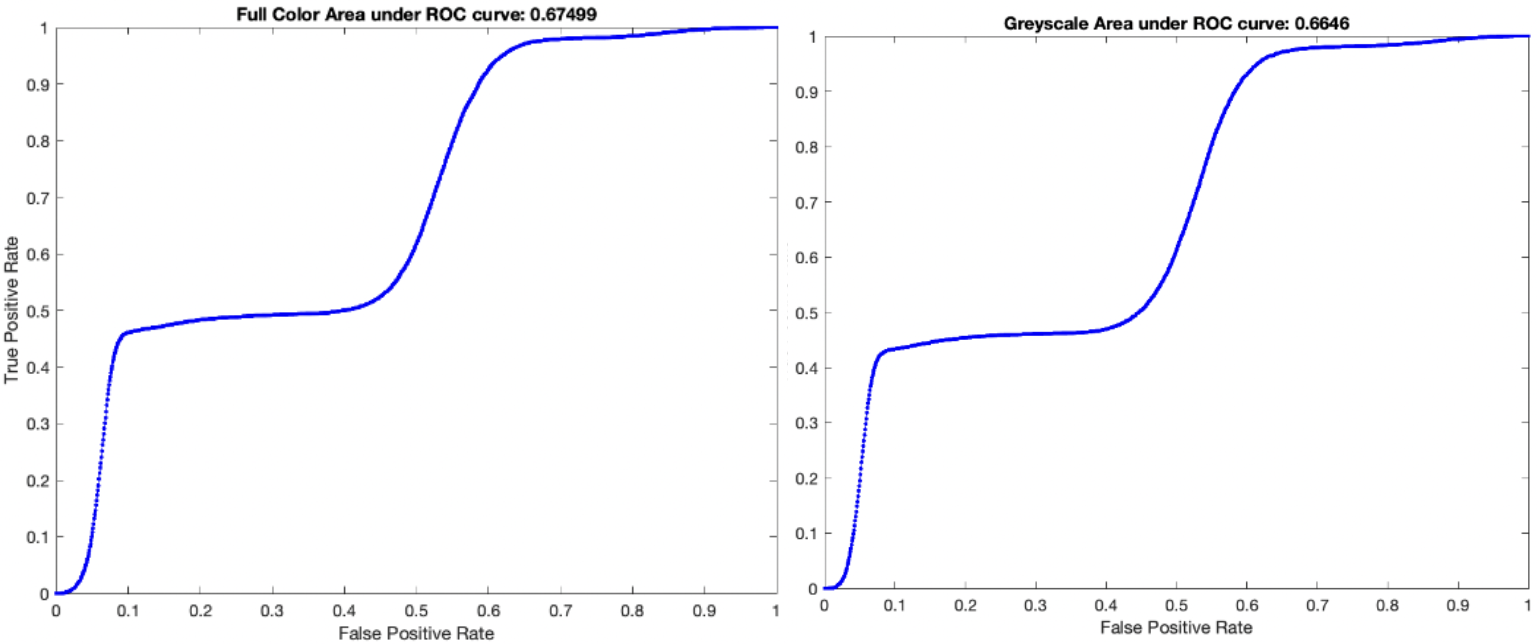}
\caption{ROC and AUC curves when using image saliency of colored and grayscale images, as a model to predict the visual fixation patterns of dogs}
   \label{fig:roc_auc_curve}
\end{figure}

The ROC curves in Figure \ref{fig:roc_auc_curve} are determined using AUC-Judd. The saliency map for each fixation image, generated using Saliency Toolbox (see Figure \ref{fig:saliency-model}), was normalized to range from 0-1 and the saliency map values were then jittered. Each of the jittered values were then used as a threshold, creating the curves \cite{salMetrics_Bylinskii}. Analyses were performed on both grayscale and color saliency mapped images.

\subsection{Appendix D: Quantitative Evaluation of Fixations per Object Class}
\label{appendix_d}

\newpage

\begin{table*}[t]
    \small
    \centering
    \caption{Quantitative evaluation of object fixations computed from fixation dataset for all participants, per object class}
    \begin{tabular}{lccccc}
    \toprule[1pt]
    \textbf{Class Object} & \textbf{Time in View [$\mu\pm\sigma$]} & \textbf{Time Fixated in View
[$\mu\pm\sigma$]}& \textbf{Size in View [$\mu\pm\sigma$]} & \textbf{$\%$ Fixation Region Occupied [$\mu\pm\sigma$]}\\
    \toprule
    bench/chair & 0.033 (0.016) & 0.012 (0.011) & 1 (0.3) & 33.3 (0.311)\\
    \midrule
    bicycle & 0.024 (0.024) & 0.099 (0.089) & 3.1(1) & 44.3 (0.251)\\
    \midrule
    building & 0.878 (0.072) & 0.144 (0.041) & 14.5 (5.1) & 53.1 (0.356)\\
    \midrule
    bus & 0.008 (0.014) & 0.348 (0.229) & 18.7 (6.3) & \textbf{81.6 (0.0307)}\\
    \midrule
    car & 0.299 (0.165) & 0.064 (0.047) & 2.7 (1) & 41.6 (0.325)\\
    \midrule
    construction & 0.011 (0.01) & 0.145 (0.12) & 4.4 (2.2) & 49.4 (0.281)\\
    \midrule
    pavement  & 0.885 (0.155) & \textbf{0.381 (0.23)} & \textbf{33.6 (9.1)} & 72.4 (0.371)\\
    \midrule
   person  & 0.389 (0.221) & 0.157 (0.111) & 13.1 (5.8) & 77.0 (0.343)\\
    \midrule
   plant horizontal & 0.616 (0.135) & 0.174 (0.105) & 18.0 (6.4) & 60.2 (0.374)\\
    \midrule
   plant vertical & \textbf{0.934 (0.061)} & 0.269 (0.194) & 21.2 (6.5) & 61.1 (0.356)\\
    \midrule
   pole  & 0.168 (0.062) & 0.027 (0.026) & 2.2 (1.1) & 33.6 (0.315)\\
    \midrule
   scooter  & 0.008 (0.005) & 0.077 (NA) & 1.2 (1.2) & 15.8 (NA)\\
    \midrule
   sculpture  & 0.037 (0.016) & 0.036 (0.03) & 1.5 (1.3) & 49.6 (0.335)\\
    \midrule
   sign & 0.013 (0.009) & 0.049 (0.03) & 2.8 (2.7) & 44.7 (0.369)\\
    \midrule
   sky  & 0.838 (0.073) & 0.07 (0.086) & 7.5 (3.3) & 49.7 (0.384)\\
    \midrule
    \bottomrule[1pt]
    \end{tabular}
    \label{tab:quantitative-results}
\end{table*}

\end{document}